\documentclass[5p, times]{elsarticle}
\usepackage{amsmath,epsfig}
\usepackage[english]{babel}
\usepackage[latin1]{inputenc}
\usepackage{amsmath}
\usepackage{amsthm}
\usepackage{amssymb}
\usepackage{eurosym}
\usepackage{multirow}
\usepackage{lscape}
\usepackage{adjustbox}
\usepackage{hyphenat}
\allowhyphens
\usepackage{subfig}
\usepackage{url}
\usepackage{bbding}
\usepackage{booktabs}% http://ctan.org/pkg/booktabs
\newcommand{\tabitem}{~~\llap{\textbullet}~~}

\usepackage{natbib}
\newcommand{\shiftpoints}{3pt}

\usepackage {tikz}
\usepackage{tikz-qtree}
\usepackage{forest}
\usetikzlibrary{shadows,trees}
\usetikzlibrary{shapes,decorations}
\usetikzlibrary{positioning}
\usetikzlibrary{fit}

\usepackage{acronym}
\usepackage[normalem]{ulem}

\newacro{}{}
\newacro{}{}
\newacro{CNN}[CNN]{convolutional neural network}
\newacro{SVM}[SVM]{support vector machine}
\newacro{SGD}[SGD]{stochastic gradient descent}
\newacro{CE}[CE]{classification error}
\newacro{LCSE}[LCSE]{level-aware context-sensitive error}

\begin{document}

\begin{frontmatter}

\title{Human experts vs. machines in taxa recognition}

\author{Johanna Ärje$^{a,b}$(\Envelope), Jenni Raitoharju$^{b}$, Alexandros Iosifidis$^{c}$, Ville Tirronen$^{d}$, Kristian Meissner$^{e}$, Moncef Gabbouj$^{b}$, Serkan Kiranyaz$^{f}$, Salme Kärkkäinen$^{a}$}
\address{$^{a}$ Department of Mathematics and Statistics, University of Jyvaskyla, P.O. Box 35 (MaD), FI-40014 University of Jyv{\"a}skyl{\"a}, Finland, johanna.arje@gmail.com\\
$^{b}$ Unit of Computing Sciences, Tampere University, Korkeakoulunkatu 1, FI-33720 Tampere, Finland\\
$^{c}$ Department of Engineering, Aarhus University, Inge Lehmanns Gade 10, DK-8000, Aarhus C, Denmark\\
$^{d}$ Faculty of Information Technology, University of Jyvaskyla, P.O. Box 35, FI-40014 University of Jyv{\"a}skyl{\"a}, Finland\\
$^{e}$ Programme for Environmental Information, Finnish Environment Institute, Survontie 9A, 40500 Jyv{\"a}skyl{\"a}, Finland\\
$^{f}$ Department of Electrical Engineering, Qatar University, Doha, Qatar}

%\author{\IEEEauthorblockN{Johanna {\"A}rje\IEEEauthorrefmark{1}\IEEEauthorrefmark{3}\IEEEauthorrefmark{7}, Jenni Raitoharju\IEEEauthorrefmark{1},
%Alexandros Iosifidis\IEEEauthorrefmark{6}, Ville Tirronen\IEEEauthorrefmark{4}, Kristian Meissner\IEEEauthorrefmark{2},
%Moncef Gabbouj\IEEEauthorrefmark{1}, Serkan Kiranyaz\IEEEauthorrefmark{5}, Salme K{\"a}rkk{\"a}inen\IEEEauthorrefmark{3}
%\\ 
%}
%\IEEEauthorblockA{
%\IEEEauthorrefmark{3}Department of Mathematics and Statistics, University of Jyv{\"a}skyl{\"a},  Jyv{\"a}skyl{\"a}, Finland\\
%\IEEEauthorrefmark{1}Department of Computing Sciences, Tampere University, Tampere, Finland\\
%\IEEEauthorrefmark{6}Department of Engineering, Aarhus University, Aarhus, Denmark \\
%\IEEEauthorrefmark{5}Department of Electrical Engineering, Qatar University, Doha, Qatar \\
%\IEEEauthorrefmark{4}Faculty of Information Technology, University of Jyv{\"a}skyl{\"a}, Jyv{\"a}skyl{\"a}, Finland\\
%\IEEEauthorrefmark{2}Programme for Environmental Information, Finnish Environment Institute, Jyv{\"a}skyl{\"a}, Finland\\
%\IEEEauthorrefmark{7}Corresponding author: johanna.arje@jyu.fi, Department of Mathematics and Statistics, P.O. Box 35 (MaD), FI-40014 University of Jyvaskyla\\
%}}

\begin{abstract}
The step of expert taxa recognition currently slows down the response time of many bioassessments. Shifting to quicker and cheaper state-of-the-art machine learning approaches is still met with expert scepticism towards the ability and logic of machines. In our study, we investigate both the differences in accuracy and in the identification logic of taxonomic experts and machines. We propose a systematic approach utilizing deep Convolutional Neural Nets with the transfer learning paradigm and extensively evaluate it over a multi-pose taxonomic dataset with hierarchical labels specifically created for this comparison.  We also study the prediction accuracy on different ranks of taxonomic hierarchy in detail. We used support vector machine classifier as a benchmark. Our results revealed that human experts using actual specimens yield the lowest classification error ($\overline{CE}=6.1\%$). However, a much faster, automated approach using deep Convolutional Neural Nets comes close to human accuracy ($\overline{CE}=11.4\%$) when a typical flat classification approach is used. Contrary to previous findings in the literature, we find that for machines following a typical flat classification approach commonly used in machine learning performs better than forcing machines to adopt a hierarchical, local per parent node approach used by human taxonomic experts ($\overline{CE}=13.8\%$). Finally, we publicly share our unique dataset to serve as a public benchmark dataset in this field.
\end{abstract}


\end{frontmatter}

\section{Introduction}

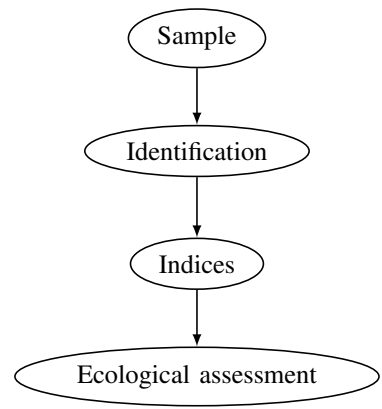
\begin{figure}[!h]
\centering
\begin {tikzpicture}[-latex ,auto ,node distance =1.5cm and 2cm ,on grid ,
semithick ,
state/.style ={ ellipse ,top color =white , bottom color = white ,
draw,black , text=black , minimum width =0.1 cm}]
\node[state] (A) {Sample};
\node[state] (B) [below=of A] {Identification};
\node[state] (C) [below=of B] {Indices};
\node[state] (D) [below=of C] {Ecological assessment};
\path (A) edge [] node[above] {} (B);
\path (B) edge [] node[above] {} (C);
\path (C) edge [] node[above] {} (D);
\end{tikzpicture}%
\caption{A schematic of the biomonitoring process.}
\label{proc}
\end{figure}

Due to its inherent slowness, traditional manual identification has long been a bottleneck in bioassessments (Fig. \ref{proc}). The growing demand for biological monitoring and the declining funding and number of taxonomic experts is forcing ecologists to search for alternatives for the cost intensive and time consuming manual identification of monitoring samples \citep{borja2013, nygard2016}. Identification of taxonomic groups in biomonitoring of, e.g., aquatic environments often involves a large number of samples, specimens in a sample, and the number of taxonomic groups to identify. For example, even in relatively species-poor regions like Finland, the calculation of the EU Water Framework Directive related indices often involves hundreds of individual specimens from 118-349 lotic diatom taxa and 44-113 lotic benthic macroinvertebrate taxa \citep{arje2016}.

While a growing body of work has used different genetic tools \citep[e.g.][]{elbrecht2017, zimmermann2015} for species identification, these methods are not yet standardized or capable of producing reliable abundance data currently required in, e.g., Water Framework Directive. While we have also worked on genetic approaches and acknowledge the great promise that genetic taxa identification methods hold \citep[e.g.][]{hering2018}, we will not explore them here but alternatively examine the suitability of machine learning techniques on image data for routine taxa identification.

Many studies on automatic classification of biological image data have been published during the past decade. \citet{kalafi2018} have done an extensive review on automatic species identification and automated imaging systems. Classification methods for aquatic macroinvertebrates have been proposed in several studies \citep[e.g.][]{culverhouse2006, lytle2010, kiranyaz2011, arje2013, joutsijoki2014, raitoharju2018}. The most popular classification methods used for identification of biological image data, such as insects, are deep neural networks and support vector machines \citep{kho2017} which are also applied in this work.

Despite the potential of computational, as well as DNA methods for taxa identification, some taxonomists continue to object the shift from manual to novel identification methods \citep{kelly2015,leese}. Often biologists that take a cursory look at automated identification tend to mistrust computational methods because they observe that a classifier is unable to separate two specimens which to them are clearly different to the human eye. Similarly, experts are baffled when the same classifier is able to discriminate between two specimens from low-resolution images while they as taxonomic experts cannot.  This mismatch in the ability of computers to identify taxa observed for single cases is often mistakenly extrapolated into an overall unreliability of algorithms. But how different truly is both the logic used and the overall accuracy of taxonomic experts and algorithms? 

Only few studies assess the accuracy of human experts and automatic classifiers, and their consequences on aquatic biomonitoring. In a study on human accuracy, \citet{haase2010} reported on the audit of macroinvertebrate samples from an EU Water Framework Directive monitoring program. They found a great discrepancy between the experts determining the true taxonomic classes and the audited laboratory workers. Contrastingly, in a study on the effect of mistakes made in automated taxa identification on biological indices, \citet{arje2017} found a relatively small impact. Literature on direct human versus machine comparisons in classification tasks in an aquatic biomonitoring context is equally scant and ambiguous. \citet{culverhouse2003} compared human and machine identification of six phytoplankton species using images and noted a similar average performance for both the experts and a computer algorithm. In \citet{lytle2010}, automatic classifiers outperformed 26 humans (a mix of experts and amateurs) when distinguishing between two stonefly taxa. Given these contrasting results, we feel it is necessary to simultaneously examine the effect of taxonomic hierarchy and of using human logical pathways for human and computer-based identification. 

Taxonomic experts identify specimens based on a predefined taxonomic resolution while automatic classifiers operate on the information of taxonomic rank used in the training data. There are different ways for accounting for data hierarchy, such as taxonomy, in classification. Hierarchical classification is widely investigated in the current literature. \citet{silla2011} sought to describe and unify the concepts of methods used in hierarchical classification problems from different domains. Using the existing literature, they categorized the classification approaches into: 1) flat classification, where the classification is performed at the most specific (deepest) rank of the taxonomy which may not always be species level, 2) local classification per level, per node or per parent node, and 3) global classification, where the whole hierarchical structure of taxonomy is taken into account at once. They found that the existing literature suggested any local or global hierarchical classifier performed better than a flat classifier, if the performance measure was specifically designed for a hierarchical structure.       

Several subsequent studies have compared flat classifiers to hierarchical classifiers. \citet{rodrigues2012} did not find a significant difference between flat and hierarchical approaches in classification of points-of-interest for land-use analysis whereas \citet{levatic2015} found that the use of hierarchy and multi-label structure improved classification results when compared to single-label cases. \citet{babbar2016} performed a theoretical study on the difference between flat and hierarchical classification and found that for well-balanced data flat classifiers should be preferred, whereas hierarchical classifiers are a better for unbalanced data.

Automatic classification of benthic macroinvertebrates, as well as plankton, has received increasing attention in recent years. However, most of the previous studies have focused on single-image data \citep[see e.g.][]{arje2010, kiranyaz2011, arje2013, joutsijoki2014, uusitalo2016, lee2016, arje2017} and have not taken the inherent hierarchical structure of the data into account. In single-image data studies, the posture of the specimens can have substantial impact on the classification. Besides \citet{lytle2010}, an imaging system producing multiple-image data is presented in \citet{raitoharju2018}. In this paper, we present a comparison of taxonomic experts and automatic classification methods on a benthic macroinvertebrate data that incorporates information on the taxonomic resolution. We test flat classifiers, local per level classifiers, and hierarchical top-down classification, i.e., local classification per parent node, and perform the automatic classification using \acp{CNN} and \acp{SVM}. The results are compared with the results of a proficiency test organized for human taxonomic experts and with a test where taxonomic experts used the same images as the automatic classifiers. The comparisons evaluate traditional single level accuracy and additionally use a novel variant of an accuracy measure that accounts for the hierarchical structure of the data.

\section{Theory}

\subsection{Hierarchy in classification}

\citet{silla2011} unified the concepts of methods used in hierarchical classification problems, and in this section we follow their terminology.

Human experts base visual identification of, e.g., invertebrate taxa on rules defined in the \citet{zoology1999}. Therefore, human experts can be thought of as hierarchical, local per parent node classifiers (see Fig. \ref{bynode}) that first identify the order of the specimen, then the family, genus, and species. The classification task is not necessarily a single level problem as some taxa need to be identified to different taxonomic levels  (see Fig. \ref{flat}) either because of predefined rules, such as minimal taxonomic requirements, or as a function of necessity when specimens lack characteristics needed to allow for better resolution. While for some taxa, genus or family might be enough, others might require species level identification depending on what the taxa information is later used for.

\tikzset{font=\small,
level distance=1.75cm,
every node/.style=
    {top color=white,
    bottom color=white,
    rectangle,rounded corners,
    minimum height=8mm,
    draw=black,
    thick,
    align=center,
    text depth = 0pt
		}
    }

\begin{figure*}[!t]
        \centering
        \subfloat[]{
                \centering
								\begin{tikzpicture}[level 3/.style={sibling distance=4mm}]
\Tree [.\node(OA){Order A} ;
        \node(FA){Family A} ; 
        [.\node(FB){Family B} ;
            [.\node(GA){Genus A} ;
                [.\node(SA){Species A} ; ]
                [.\node(SB){Species B} ; ] ]
            \node(GB){Genus B} ; ] 
        [.\node(FC){Family C} ;
            [.\node(GC){Genus C} ;
								[.\node(SC){Species C} ; ] ] ]
]
\begin{scope}[path/.style={
rounded corners=10,
draw=black},
shifttl/.style={shift={(-\shiftpoints,\shiftpoints)}},
shifttr/.style={shift={(\shiftpoints,\shiftpoints)}},
shiftbl/.style={shift={(-\shiftpoints,-\shiftpoints)}},
shiftbr/.style={shift={(\shiftpoints,-\shiftpoints)}}]
\draw[dashed,rounded corners=10]
		 ([shifttl] FA.north west)
	-- ([shifttr] FA.north east)
	-- ([shifttr] SA.north west)
	-- ([shifttl] SB.north east)
	-- ([shifttl] GB.north west)
	-- ([shifttr] GB.north east)
	-- ([shifttr] SC.north west)
	-- ([shifttr] SC.north east)
	-- ([shiftbr] SC.south east)
	-- ([shiftbl] SC.south west)
	-- ([shiftbl] GB.south east)
	-- ([shiftbr] GB.south west)
	-- ([shiftbr] SB.south east)
	-- ([shiftbl] SA.south west)
	-- ([shiftbl] FA.south east)
	-- ([shiftbl] FA.south west)
	-- cycle;
\end{scope}
\end{tikzpicture}
\label{flat}
        }
        \subfloat[]{
                \begin{tikzpicture}[level 3/.style={sibling distance=4mm}]
\Tree [.\node(OA){Order A} ;
        \node(FA){Family A} ; 
        [.\node(FB){Family B} ;
            [.\node(GA){Genus A} ;
                [.\node(SA){Species A} ; ]
                [.\node(SB){Species B} ; ] ]
            \node(GB){Genus B} ; ] 
        [.\node(FC){Family C} ;
            [.\node(GC){Genus C} ;
								[.\node(SC){Species C} ; ] ] ]
]
\draw[dashed,rounded corners=10]($(FA) +(-0.9,0.5)$)rectangle($(FC) +(0.9,-0.5)$);
\end{tikzpicture}
\label{perlevel}
        }\\
        \subfloat[]{
                \begin{tikzpicture}[level 3/.style={sibling distance=4mm}]
\Tree [.\node(OA){Order A} ;
        \node(FA){Family A} ; 
        [.\node(FB){Family B} ;
            [.\node(GA){Genus A} ;
                [.\node(SA){Species A} ; ]
                [.\node(SB){Species B} ; ] ]
            \node(GB){Genus B} ; ] 
        [.\node(FC){Family C} ;
            [.\node(GC){Genus C} ;
								[.\node(SC){Species C} ; ] ] ]
]
\draw[dashed,rounded corners=10]($(OA) + (-0.8,0.5)$)rectangle($(OA) +(0.8,-0.5)$);
\draw[dashed,rounded corners=10]($(FA) + (-0.9,0.5)$)rectangle($(FA) +(0.9,-0.5)$);
\draw[dashed,rounded corners=10]($(FB) + (-0.9,0.5)$)rectangle($(FB) +(0.9,-0.5)$);
\draw[dashed,rounded corners=10]($(FC) + (-0.9,0.5)$)rectangle($(FC) +(0.9,-0.5)$);
\draw[dashed,rounded corners=10]($(GA) + (-0.8,0.5)$)rectangle($(GA) +(0.8,-0.5)$);
%\draw[dashed,rounded corners=10]($(GB) + (-0.8,0.5)$)rectangle($(GB) +(0.8,-0.5)$);
%\draw[dashed,rounded corners=10]($(GC) + (-0.8,0.5)$)rectangle($(GC) +(0.8,-0.5)$);
%\draw[dashed,rounded corners=10]($(SA) + (-0.9,0.5)$)rectangle($(SA) +(0.9,-0.5)$);
%\draw[dashed,rounded corners=10]($(SB) + (-0.9,0.5)$)rectangle($(SB) +(0.9,-0.5)$);
%\draw[dashed,rounded corners=10]($(SC) + (-0.9,0.5)$)rectangle($(SC) +(0.9,-0.5)$);
\end{tikzpicture}
\label{bynode}
        }
        \caption{Different types of classifiers for hierarchical data: (a) Flat classification, (b) Local classification per level, (c) Local classification per parent node. The dashed boxes represent a single trained classifier.}
				\label{hierclass}
\end{figure*}
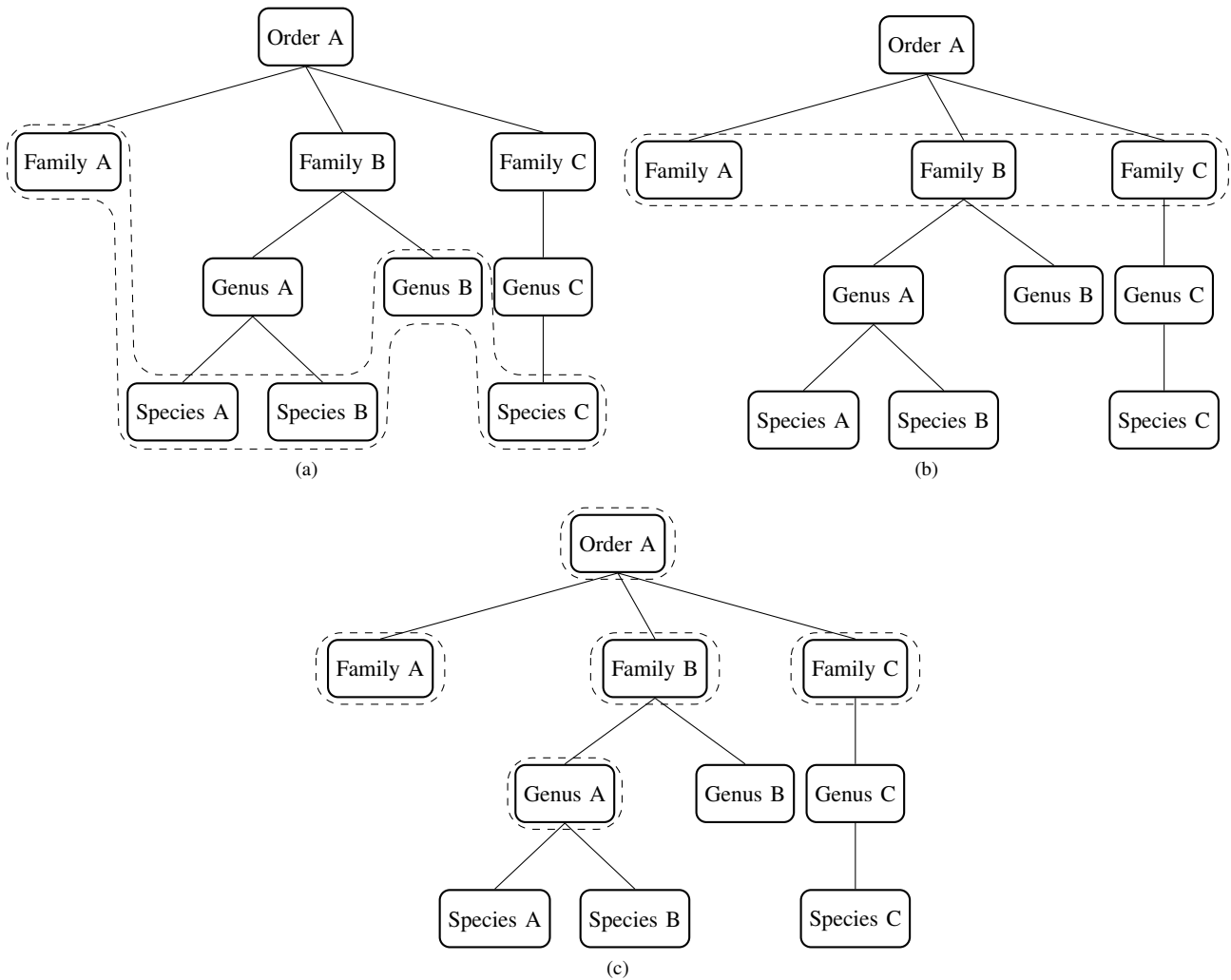

Usually, automatic classification methods have no information on the possible hierarchical nature of the data. The classifiers simply aspire to identify the specimens to the class labels provided in the training data. In the case of benthic macroinvertebrate data, the class labels represent a mix of families, genera, and species. An algorithm working this way is called a flat classifier as it is not aware that species A and B belong to the same genus A, but uses the same approach to distinguish them from each other as when separating species A from genus B. Flat classification produces a single label prediction for each specimen but the hierarchical level of that label may vary depending on the data (Fig. \ref{flat}).

Depending on what the taxa information is later used for, it could be beneficial to build a classifier that identifies a certain taxonomic rank well. For example, a common biological index used in river macroinvertebrate biomonitoring is the number of typical EPT families (\textit{Ephemeroptera}, \textit{Plecoptera}, \textit{Trichoptera}). For the purpose of evaluating this index, it would be reasonable to train a classifier to identify the family level with high accuracy. However, such a classifier trained with the family level labels would have no intrinsic information on certain families descending from the same order. This type of a classification scheme is known as local classification per level (see Fig. \ref{perlevel}). One could build a classification system with local level classifiers for each level of the hierarchy. While such a system would predict multiple labels for each specimen there would be no guarantee that the predictions for the different levels are taxonomically coherent.

It is also possible to build a hierarchical classification system that accounts for the hierarchical nature of the data and force it to operate in the same manner as human experts. This requires to build a sequence of several classifiers: i) an order level classifier to predict the order of each specimen, ii) multiple family level classifiers, one for each possible order present in the data, iii) multiple genus level classifiers, one for each family present in the data, and finally, iv) multiple species level classifiers to predict the species within each genus. This type of a hierarchical classification scheme is known as local classification per parent node and it predicts the labels for each rank of taxonomic resolution for all the specimens in the data (see Fig. \ref{bynode}). While a human-like hierarchical classifier is guaranteed to logically follow taxonomy all classification errors made on higher levels of hierarchy will propagate to the lower level predictions.

The focus of this work is on the comparison of identification results obtained by taxonomic expert logic and machine logic. As traditional machine logic uses flat classification and taxonomic expert logic can be thought of as local classification per parent node, we will not consider global hierarchical classifiers.

\subsection{Performance measures}

Traditionally, classification methods are compared based on their accuracy, which is the proportion of correct predictions, or \ac{CE},
$$CE=\frac{1}{n}\sum_{i=1}^n L(\hat{y}_i,y_i),$$
where $L(\cdot,\cdot)$ is a 0-1 loss function and $n$ is the total number of observations. Other measures of performance such as false positive rate, false negative rate, sensitivity, and specificity can also be calculated from the confusion matrix and take single label predictions into account. These performance measures can be calculated for both flat classification (Fig. \ref{flat}) or for each level of local classification (Fig. \ref{perlevel}, \ref{bynode}).

With hierarchical data, each observation has multiple labels and we need to measure the performance as a whole accounting for all the labels. \citet{verma2012} presented context sensitivite loss (CSL) function which takes the top-down success into account. They used this loss function to define context-sensitive error (CSE),

\begin{eqnarray*}
CSE&=&\frac{1}{nH}\sum_{i=1}^n L(\hat{y}_i,y_i),
\end{eqnarray*}
\noindent where
\begin{eqnarray*}
L(\hat{y}_i,y_i)&=&\left\{\begin{array}{l l l} h \text{, where $h$ is the height of the deepest}\\
													\,\,\,\,\text{common ancestor of pair }  (\hat{y}_i,y_i)\\
												 0,\text{ if } \hat{y}_i=y_i\end{array}\right.\\
\end{eqnarray*}
\noindent and $H$ is the total number of levels in the hierarchy.

Because the deepest available level of hierarchy can vary in taxonomic data, we propose to modify the measure to a \ac{LCSE},
\begin{eqnarray*}
LCSE&=&\frac{1}{n}\sum_{i=1}^n \frac{1}{H_i}L(\hat{y}_i,y_i),
\end{eqnarray*}
\noindent where $L(\hat{y}_i,y_i)$ is as above and $H_i$ is the number of available levels in the hierarchy for observation $i$.

\section{Materials and methods}

\subsection{Proficiency test for human experts}\label{humans}

In order to compare automatic and manual classification, we needed classification results on the same set of taxa for both. The Finnish Environment Institute (SYKE), an appointed National Reference Laboratory in the environmental sector in Finland, organized a proficiency test on taxonomic identification of boreal freshwater lotic, lentic, profundal, and North-Eastern Baltic benthic macroinvertebrates in 2016. The aim of the test was to assess the reliability of professional and semi-professional identification of macroinvertebrate taxa routinely encountered during North-Eastern Baltic coastal or boreal lake and stream monitoring \citep{meissner2017}. A part of the proficiency test included 10 participants who all identified a different set of 50 specimens of lotic freshwater macroinvertebrates belonging to a total of 46 taxonomic groups, of which 39 are in common with the multiple-image data introduced in the following Section \ref{img} (see taxa list in Table \ref{numbers}). The samples sent out to the participants included 0--4 specimens of each taxa. The class labels of the 39 overlapping taxa consisted of 26 species, 12 genera, and one family. The chosen taxonomic resolution is based on the requirements for the Finnish national freshwater monitoring program for macroinvertebrates \citep{meissner2010}. The 'true' labels of the specimens were predetermined by an expert panel and the specimens were shipped to the participants. Participants were provided with the list of the almost 300 possible taxa labels \citep{meissner2017}.

\subsection{Image data}\label{img}

\begin{figure}[!t]
\centering
\includegraphics[width=8cm]{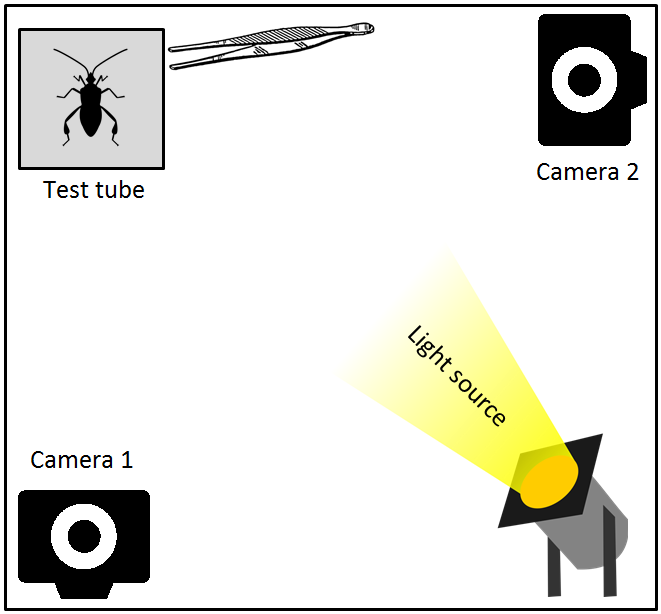}
\caption{Schematic of the imaging system for macroinvertebrates pictured from above.}
\label{imaging}
\end{figure}

\begin{figure}[!h]
\centering
\includegraphics[width=8cm]{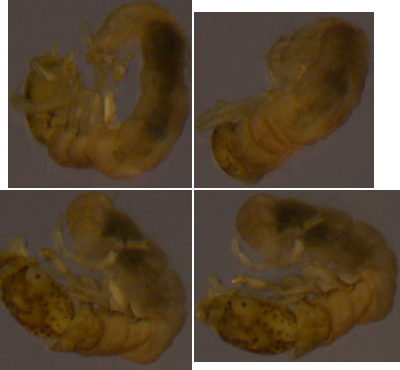}
\caption{Example images of a \textit{Polycentropus flavomaculatus} specimen from two cameras. The top row images are from camera 1 and the bottom row images from camera 2.}
\label{multipleimg}
\end{figure}

We produced all images with a new imaging system described in \citet{raitoharju2018} that allows for multiple images per specimen. The system is illustrated in Fig. \ref{imaging}. It consists of two Basler ACA1920-155UC cameras (frame rate of 150 fps) with Megapixel Macro Lens (f=75mm, F:3.5-CWD\(<\)535mm) placed at a 90 degree angle to each other, a high power LED light and a cuvette (i.e. a rectangular test tube) in a metal container. The device is sealed with a lid to block any extra light. The imaging system has a software that builds a model of the background of the cuvette filled with alcohol and sets off the cameras when a significant change in the view of the camera is detected. When a macroinvertebrate specimen is put into the cuvette, it sinks and both cameras take multiple shots of it (Fig. \ref{multipleimg}). The number of images per specimen depends on the size and weight of each specimen: Heavier specimens sink faster, leading to a smaller number of images. Compared to the system and data described in \citet{raitoharju2018}, we have improved the system to handle more than two images per specimen.

\begin{figure*}[!t]
\centering
\includegraphics[width=17cm]{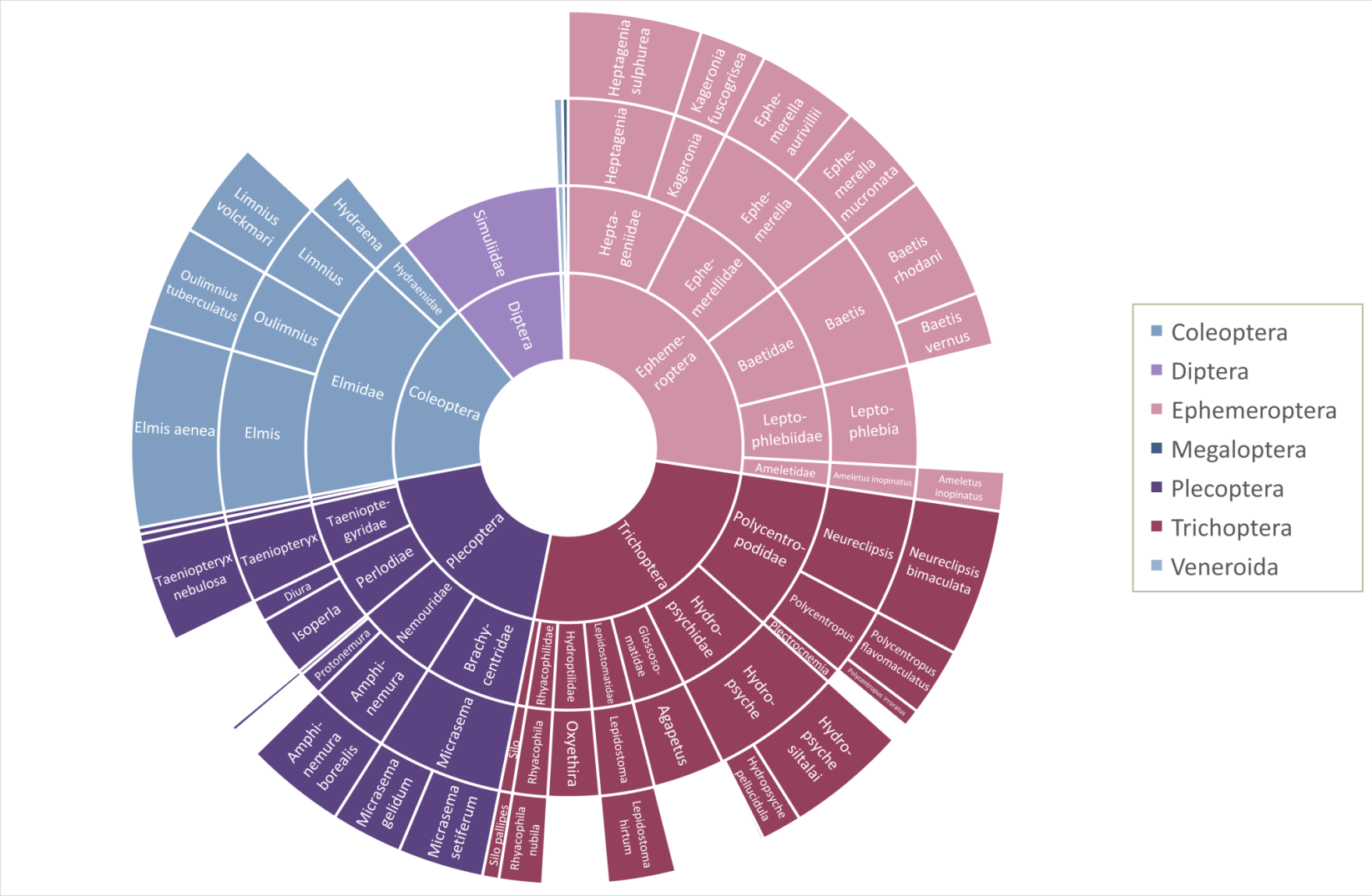}
\caption{Taxonomic resolution and distribution of the multilabel image data. The area of the slices represent the relative size of each taxonomic group at the different ranks of taxonomic hierarchy.}
\label{taxadistribution}
\end{figure*}

In Finland, the national reference taxa list sets the taxonomic ranks to which human experts are required to identify specimens monitoring samples \citep{meissner2010}. In the human proficiency test only a subset of the taxa from the national reference taxa list is used. The choice of the specific taxa and specimens used in the proficiency test is determined both by relevance of the taxa in national assessment indices, the availability of adequate testing material and to a lesser degree the inclusion of easily misidentified taxa. Human participants were required to key 50 specimens in total for the river benthic subtest \citep{meissner2017}. Using the described imaging device, the Finnish Environment Institute compiled a new image database of 126 lotic freshwater macroinvertebrate taxa and over 2.6 million images. This data has 39 taxa overlapping with those present in the human proficiency test which are used in the current work (Table \ref{numbers}). We restricted the number of images per specimen to a maximum of 50 images for computational reasons. If a specimen had more images from both cameras combined, we randomly selected 50 of them. The final data comprises 9631 observations and a total of 460004 images belonging to 39 taxa at the deepest available taxonomic rank. In total, considering one taxonomic rank at a time, the data consists of 7 orders, 23 families, 30 genera, and 26 species (see Fig. \ref{taxadistribution}). The number of specimens for each taxa and the taxonomic resolution are shown in Table \ref{numbers}. The image resolution for this data varies from $32\times 20$ pixels to $468\times 540$ pixels. The 'true' labels for the specimens were defined by a group of taxonomic experts. While we acknowledge that there might be some mislabeled specimens, combining the knowledge of multiple taxonomic experts should improve the accuracy \citep{caley2014}. We provide the data for public use in \url{http://urn.fi/urn:nbn:fi:att:dc6440ad-43bd-4349-8fb9-0e0d1971a7e8}.

\subsection{Classification set-up}

To have classification results comparable to the proficiency test, we compiled a set of data divisions for the image data with the exact same number of test specimens per taxa as in the proficiency test. As the proficiency test had 10 participants identifying lotic freshwater macroinvertebrates, we created 10 data divisions. The test sets comprise randomly selected 45--46 specimens belonging to the 39 taxonomic groups present in both the physical data and the image data. The test sets have an approximately equal number of specimens from each class. We divided the rest of the specimens of each data split for training (80 \%) and validation (20 \%). Due to the nature of the collected data, the training and validation data are unbalanced. In the following sections, these data sets are referred to as the ''comparison data''. The number of specimens per test set in the comparison data is lower than in the proficiency test because 4--5 specimens sent to each participant belonged to taxonomic groups not present in the image data.

Since the comparison between professionals and semi-pro-fessionals analysing physical data with a laboratory microscope and automatic classifiers using image data is unequal, we asked the proficiency test participants to also try to identify taxa from the test images of the comparison data. Each participant received one of the test sets and a list of the 39 possible taxa labels. To avoid fatigue and to encourage more experts to participate, we restricted the number of images per test specimen to 10. The automatic classifiers used exactly the same test data. In addition, because some of the images are fuzzy, the experts were allowed to classify the taxa to a higher taxonomic rank if they were unsure. The automatic classifiers always predicted the classes of the test specimens to the deepest available rank of taxonomic resolution. Of the ten experts participating in the proficiency test, three volunteered to take part in this image classification study.

As the comparison test sets are very small, we also studied the performance of the automatic classifiers on larger test sets. We split the specimens randomly into training (70 \%), validation (10 \%) and test (20 \%) data 10 times. This time the number of specimens in each taxon varied in all training, validation, and test sets depending on the size of the taxa in the dataset. We refer to these sets as the ''machine learning data'' as the splitting is typical for machine learning, but not suitable for comparisons with humans. For the test sets in the machine learning data, we included all images (max. 50) per specimen.

We considered different approaches to take the hierarchical nature of the data into account: A flat classifier is a single classifier with the 39 taxa as output labels. Local per level classifiers are built for each taxonomic rank separately: a classifier for the orders and another classifier for the families. We only trained local per level classifiers for the two highest taxonomic ranks as some of the taxa in the data have information only on these ranks. The top-down, local per parent node classifier is a system comprising 17 classifiers: one classifier at the top to identify the order of a specimen, four classifiers at the family level as there are four families with more than one genus within them, five classifiers at the genus level, and seven classifiers at the species level (see Table \ref{numbers}). Some of the specimens get their predictions already at the order level since there are three orders with only one family or genus within them. In the data, there are two genera (\textit{Leuctra sp.} and \textit{Nemoura sp.}) for which only some of the specimens have information on species (\textit{Leuctra nigra} and \textit{Nemoura cinerea}). To separate these groups with the local per parent node classification approach, we temporarily marked the species for the rest of the \textit{Leuctra sp.} and \textit{Nemoura sp.} specimens as '$0$'. We trained the local species level classifiers and if they predicted the '$0$' label, we marked the specimen as predicted only to genus level.

\subsection{Classification methods}

We selected our methods for the automatic classification to be \ac{CNN} \citep{krizhevsky2012} and \ac{SVM} \citep{cortes1995} which are the most popular ones used for biological image data \citep{kho2017}. As our \ac{CNN} model, we used the MatConvNet \citep{matconvnet} implementation of the AlexNet \ac{CNN} architecture \citep{krizhevsky2012}. The architecture has five convolution layers followed by three fully-connected layers. The last fully-connected layer is followed by a softmaxloss(train)/softmax(test) layer. In our tests, we considered also the output of the last fully-connected layer instead of the softmax output, because we observed that this produced better results, when the final class was decided based on the average of the outputs for each image of a specimen. We trained flat and local per level classifiers from scratch using 60 training epochs. For the 17 classifiers of each local per parent node classifier, we took the flat classifier for the corresponding data split as our starting point and fine-tuned the network for 10 epochs (5 epochs only the last fully-connected layer, 3 epochs all fully-connected layers, and 2 epochs all layers). In all cases, we used a batch size of 256 and trained the network using stochastic gradient descent with a momentum of 0.9. When training from scratch, we used a learning rate varying from 0.01 to 0.0001 and for fine-tuning a learning rate varying from 0.005 to 0.0001. We saved the networks after each epoch and selected the final model based on the classification accuracy on the validation set.

While \acp{CNN} use the original images as input, we extracted a set of 64 simple geometry and intensity-based features from the images using ImageJ \citep{Rasband} for \acp{SVM}. The geometric features extracted include, e.g., area, perimeter, width and height of a bounding rectangle, while the intensity-based features were extracted from gray, red, green, and blue scale channels of the images. The complete set of features is listed in detail in \ref{features}. As these features are simple and the classification task of identifying such a large number of classes is a complex one, we found that making a principal component transformation on the features improves classification results. Therefore, we performed a principal component transformation, as well as standardization, on the features before using them for classification.

We built our \ac{SVM} model \citep{chang2011} using R \citep{R} package e1071 \citep{e1071} and used a Gaussian kernel. For flat classification and local per level classification, we performed a grid search for the parameters over $c=\{2^8,2^9,2^{10},2^{11}\}$ and $\gamma=\{2^{-11},2^{-10},2^{-9},2^{-8}\}$. For the local per parent node hierarchical classification system, we explored a larger grid as the classification problems can be very different from another at different nodes of the hierarchical system. Due to the amount of data and time consumed by evaluating just a single parameter combination, we did the following: we randomly selected one image per specimen and used this data to perform the grid search for the parameters over $c=\{2^1,2^2,\ldots,2^{15}\}$ and $\gamma=\{2^{-15},2^{-14},\ldots,2^{-1}\}$. After determining the optimal parameter values with this smaller data, we did a small, $3\times3$, grid search around those values with all the images (max. 50 images per specimen).

For both, the comparison and the machine learning data, we did the following: With each data split, we used the training data to train the model and the validation data to either select the best epoch to stop training (\acp{CNN}) or select optimal parameter values (\acp{SVM}) based on the classification accuracy of the validation specimens. With \acp{SVM}, we combined the training and validation data to train the final model after fixing the parameters. At the end, we classified each test image and selected the final class for each specimen using either average output (\acp{CNN}) or majority vote over all the images of the specimen (\acp{CNN}, \acp{SVM}).

\section{Analysis and inference}

\subsection{Comparison data}\label{balanced_res}

Classification results for the comparison test sets of the image data as well as results of the proficiency test on physical data are presented in Table \ref{balanced}. The first row of results shows the average \ac{CE} on the deepest available rank of taxonomy. These are the results traditionally examined with flat classifiers. Taxonomic experts using physical data and microscopes to identify the taxa still outperform the automatic approaches. This result by taxonomic experts can be considered as a gold-standard to compare to. However, taxonomic experts predicting taxa from the images make the most classification errors. This is understandable as the image quality can be sub-par for some specimens and the experts have not studied identification from these types of images. For the automatic classifiers, \ac{CNN} using the flat classification approach and the average output for deciding the final class has the lowest CE and is in the range of taxonomic experts with physical data. The average output clearly outperforms the majority vote as a decision rule for the final class even though the number of images per specimen is relatively high.

\begin{table*}[htbp]
\begin{center}
\begin{tabular}{|l|l|c|c|c|c|c|c|c|c|c|}
\hline
&	&	\bf{\ac{CNN}}&	\bf{\ac{CNN}}&	\bf{\ac{CNN}}&	\bf{\ac{CNN}}&	\bf{\ac{SVM}}&	\bf{\ac{SVM}}&	\bf{\ac{SVM}}&	\bf{Experts}&	\bf{Experts}\\
&	&	flat,&	flat,&	local/&	hier. &	flat&	local/ &	hier. &	images&	physical\\
&	&	aver.&	vote&	level&	&	 &	level &	&	&	data\\
\hline										
\multirow{4}{*}{Deepest level}&	$\overline{CE}$&	0.114&	0.131&	&	0.138&	0.243&	&	0.28&	0.553&	0.061\\
&	$sd(CE)$&	0.036&	0.054&	&	0.055&	0.081&	&	0.074&	0.153&	0.053\\
&	$\overline{\ac{LCSE}}$&	0.052&	0.070&	&	0.070&	0.173&	&	0.191&	0.353&	0.028\\
&	$sd(\ac{LCSE})$&	0.023&	0.034&	&	0.036&	0.061&	&	0.053&	0.162&	0.024\\
\hline										
\multirow{2}{*}{Order}&	$\overline{CE}$&	0.004&	0.018&	0.011&	0.011&	0.085&	0.075&	0.075&	0.210&	0.007\\
&	$sd(CE)$&	0.009&	0.02&	0.012&	0.012&	0.041&	0.026&	0.026&	0.190&	0.015\\
\multirow{2}{*}{Family}&	$\overline{CE}$&	0.039&	0.059&	0.150&	0.059&	0.173&	0.181&	0.193&	0.291&	0.020\\
&	$sd(CE)$&	0.029&	0.037&	0.259&	0.044&	0.070&	0.062&	0.069&	0.151&	0.020\\
\hline										
\multirow{4}{*}{Error structure}&	\#ERR(order) &	2&	8&	&	5&	39&	&	34&	29&	3\\
&	\#ERR(family) &	16&	19&	&	22&	40&	&	54&	11&	6\\
&	\#ERR(genus) &	12&	12&	&	12&	16&	&	22&	15&	6\\
&	\#ERR(species) &	22&	21&	&	24&	16&	&	18&	21&	13\\
\hline
\end{tabular}
\caption{Classification results for comparison test data. \ac{CE} and \ac{LCSE} are averaged over all 10 experts/data splits (for experts with images, 3 data splits). The number of new classification errors at each taxonomic rank is summed over all 10 data splits, where $n_{total}=457$ (for experts with images, 3 data splits, $n_{total}=137$).}
\label{balanced}
\end{center}
\end{table*}

While flat classification gives only a single level and single label predictions, it is still possible to make comparisons on different ranks of taxonomic resolution. We simply take the predictions from the deepest rank of taxonomy of the data and add the ascending taxa labels accordingly. Let us call this a bottom-up examination. Using the bottom-up examination, we can calculate \ac{LCSE} also for flat classifiers. The \ac{LCSE} values for all classifiers as well as for taxonomic experts are clearly smaller than the \ac{CE} values (see Table \ref{balanced}). This means that most of the classification errors occur on deeper ranks of taxonomic resolution while the order and family might be predicted correctly. If all the classification errors were done already on the order level, \ac{CE} and \ac{LCSE} would be the same. For taxonomic experts using physical data, \ac{LCSE} is close to zero as expected since taxonomic experts use a top-down hierarchical logic for the classification task, and identifying the higher ranks of taxonomy should be an easy task for an expert. Also in terms of \ac{LCSE}, \acp{CNN} get close to the taxonomic expert level.

Contrary to the previous findings in hierarchical classification literature \citep{silla2011}, the flat classifiers for both CNN and SVM produce better results than the hierarchical classification approach. \citet{babbar2016} stated in their study that if the data is highly unbalanced, hierarchical classifiers are better options even though their empirical error (\ac{CE}) may be higher due to error propagation. While our test data is balanced, the training data used to train the classifiers is not. However, taking the hierarchical nature of the data into account when building the classifier produces not only a higher \ac{CE} but also a little higher \ac{LCSE}. It is worth noting that the optimization of the classifiers is based on CE, not \ac{LCSE}. The only improvement the hierarchical classification system offers is a slightly lower \ac{CE} on the order level for SVM. Note that for the order level, the hierarchical classifier and the local per level classifier are the same. Interestingly, the local per level \ac{SVM} and \ac{CNN} classifiers for family level perform worse than the flat classifiers with the ascending taxa labels. The notably high \ac{CE} for local per level \ac{CNN} for family level is due to data split three, where \ac{CNN} classifies all observations to the family \textit{Elmidae}. When leaving this data split out, the average classification error is 7 \%.

The bottom part of Table \ref{balanced} shows the error structure for each classifier and the taxonomic experts. The number of new errors at the different taxonomic ranks sum up to the total amount of misclassifications for the 10 balanced test splits. The difference in taxonomic expert and machine logic is evident through the number of errors on each taxonomic rank. For taxonomic experts using physical data, there are very few misclassifications at the order level and the number of errors increases with the taxonomic resolution. For experts using image data, all the order level errors are due to completely missing predictions for images being too challenging to identify. That is, all the predictions made by the experts were correct at the order level and as with physical data, the number of errors increases as with the taxonomic rank. For the automatic classifiers, most misclassifications are made at either species or family level. There is no such clear hierarchy in the error structure as for the taxonomic experts.

In biomonitoring and ecosystem assessment, not only a low number of classification errors is essential, but also the type of errors made as some misclassifications can have higher cost than others. To examine this, we analysed the confusion matrices of the classifiers and taxonomic experts. Concerning especially demanding taxa, both the taxonomists and automatic classifiers had difficulties identifying \textit{Hydropsyche saxonica}. Human experts easily misclassified them as \textit{Hydropsyche angustipennis} when using physical data and into a mix of other \textit{Hydropsyche} species when using image data. The image data has no \textit{Hydropsyche angustipennis} specimens and the automatic classifiers predicted many of the \textit{Hydropsyche saxonica} to be \textit{Hydropsyche pellucidula} (see Fig. \ref{ssbd}). \textit{Hydropsyche saxonica} is also one of the least represented taxa in the image data with only 17 specimens (see Table \ref{numbers}) which is likely to be the reason the automatic classifiers have trouble classifying them. Besides this taxa, the human experts had another challenging taxa in the physical data. Some \textit{Rhyacophila nubila} were misclassified as \textit{Rhyacophila fasciata}. With the more difficult image data, the taxonomic experts classified these individuals to genus level only or left them unidentified, while \acp{SVM} mixed them with other taxa as there were no \textit{Rhyacophila fasciata} in the image data. In addition, with the image data, the human experts had trouble identifying the \textit{Coleopteran} \textit{Elmis aenea} with some of them unidentified completely and some of them misclassified as the \textit{Coleopteran} \textit{Oulimnius tuberculatus}. The automatic classifiers identified this taxon more easily. 

\subsection{Machine learning data}

The results on the machine learning data with larger test sets are shown in Table \ref{unbalanced}. Both \ac{CE} and \ac{LCSE} for all the classifiers are clearly lower with these data splits. That is due to two factors: these results are more stable, meaning they are not affected by individual difficult specimens, and here the size of each taxa in the test set reflects the size of the taxa in the training/validation sets. The comparison test sets of Section \ref{balanced_res} had only 0--4 specimens of each taxa and therefore the taxa with only few training specimens had the same weight as the taxa with hundreds of training specimens. For the machine learning data, taxa with little training data will also have only few test specimens and a small weight on the classification error of the entire test set.

\begin{table*}[htbp]
\begin{center}
\begin{tabular}{|l|l|c|c|c|c|c|c|c|}
\hline						
&	&	\bf{\ac{CNN}}&	\bf{\ac{CNN}}&	\bf{\ac{CNN}}&	\bf{\ac{CNN}}&	\bf{\ac{SVM}}&	\bf{\ac{SVM}} &	\bf{\ac{SVM}}\\
&	&	flat,&	flat,&	local/&	hier. &	flat&	local/ &	hier. \\
&	&	aver.&	vote&	level&	&	 &	level &	\\			
\hline										
\multirow{4}{*}{Deepest level}&	$\overline{CE}$&	0.078&	0.087&	&	0.087&	0.17&	&	0.181\\
&	$sd(CE)$&	0.009&	0.009&	&	0.013&	0.008&	&	0.009\\
&	$\overline{\ac{LCSE}}$&	0.044&	0.052&	&	0.048&	0.124&	&	0.129\\
&	$sd(\ac{LCSE})$&	0.006&	0.006&	&	0.005&	0.006&	&	0.008\\
\hline								
\multirow{2}{*}{Order}&	$\overline{CE}$&	0.01&	0.015&	0.011&	0.011&	0.055&	0.053&	0.053\\
&	$sd(CE)$&	0.002&	0.003&	0.002&	0.002&	0.006&	0.005&	0.005\\
\multirow{2}{*}{Family}&	$\overline{CE}$&	0.041&	0.05&	0.033&	0.044&	0.129&	0.126&	0.135\\
&	$sd(CE)$&	0.006&	0.007&	0.003&	0.004&	0.006&	0.008&	0.011\\
\hline								
\multirow{4}{*}{Error structure}&	\#ERR(order) &	194&	287&	&	216&	1071&	&	1017\\
&	\#ERR(family) &	605&	685&	&	638&	1428&	&	1589\\
&	\#ERR(genus) &	304&	307&	&	319&	455&	&	505\\
&	\#ERR(species) &	410&	412&	&	510&	344&	&	393\\
\hline
\end{tabular}
\caption{Classification results for machine learning test data. \ac{CE} and \ac{LCSE} are averaged over all 10 data splits, where each test split has $n=1937$. The number of new classification errors at each taxonomic rank is summed over all 10 data splits, where $n_{total}=19370$.}
\label{unbalanced}
\end{center}
\end{table*}

The results are similar to those in Table \ref{balanced}. \acp{CNN} produce the best classification results. Again, the flat classification versions of \ac{CNN} and \ac{SVM} outperform the hierarchical classifiers contradicting previous findings of hierarchical classification studies \citep{silla2011}. With the machine learning data splits, the local per level classification approach gives slightly lower \ac{CE} than the flat classifier on both order level (\ac{SVM}) and family level (\ac{SVM} and \ac{CNN}).

When considering individual challenging taxa, the best classifier, \ac{CNN}, has mostly trouble with the least represented taxa in the data due to lack of adequate training data. The smallest taxa are \textit{Hydropsyche saxonica}, \textit{Nemoura cinerea}, \textit{Capnosis schilleri}, \textit{Sialis sp.}, \textit{Leuctra nigra} and \textit{Sphaerium sp.} with average number of specimens in the training data, $N=\{13, 11, 15,$ $19, 19, 107\}$ and $\#\text{images}=\{349, 540, 730, 856, 960, 1239\}$ respectively. With the exceptions of \textit{Sialis sp.} and \textit{Sphaerium sp.}, the average \ac{CE} for these taxa ranged from 62\% to 98\% for \acp{CNN} and from 61\% to 100\% for \acp{SVM}. On the contrary, all the classifiers performed well on classifying \textit{Sphaerium sp.} ($\overline{CE}\in \left[0,8\%\right]$), and \acp{CNN} also relatively well on classifying \textit{Sialis sp.} ($\overline{CE}\in \left[15,18\%\right]$).

\begin{figure}[!t]
\centering
\includegraphics[width=8cm]{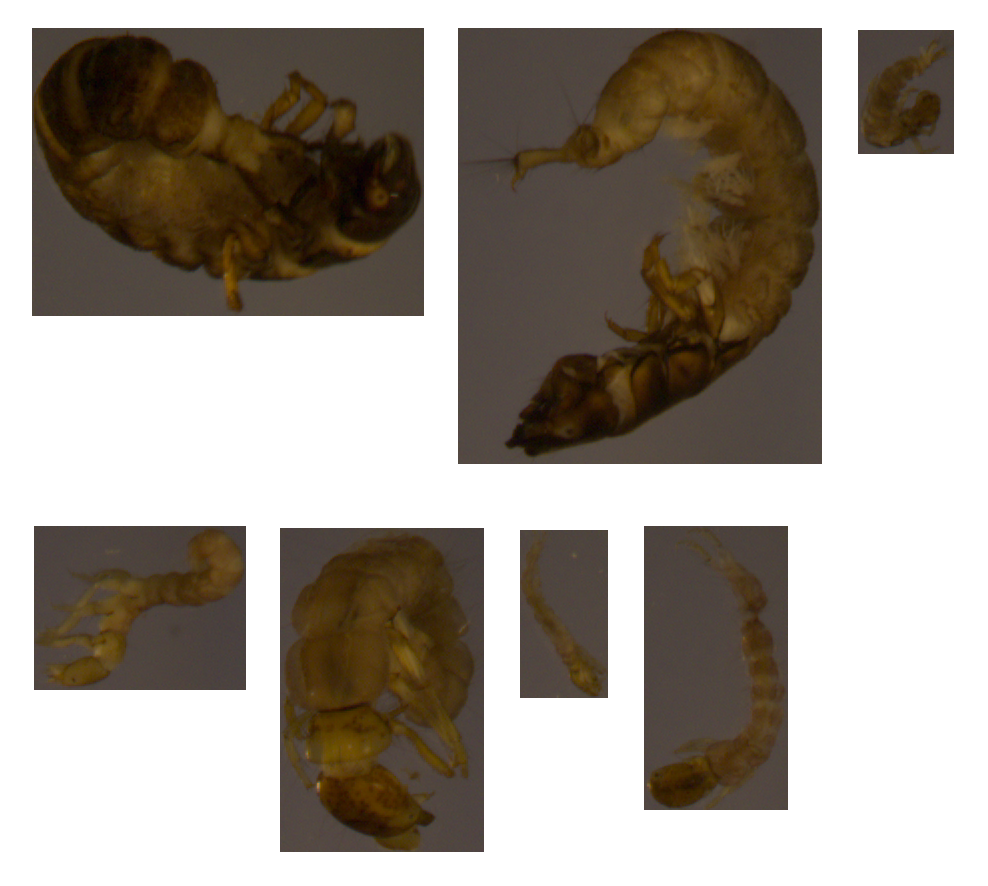}
\caption{Examples of visual differences among taxa belonging to the same family or genus. Top row: \textit{Hydropsyche pellucidula}, \textit{Hydropsyche saxonica}, and \textit{Hydropsyche siltalai} all belong to the genus \textit{Hydropsyche sp}. Bottom row: \textit{Neureclipsis bimaculata}, \textit{Plectronemia}, \textit{Polycentropus flavomaculatus}, and \textit{Polycentropus irroratus} all belong to the family \textit{Polycentropodidae}. In both cases, the taxa are of different sizes and colors.}
\label{ssbd}
\end{figure}

One reason why the hierarchical, local per parent node approach performs worse than flat classification could be that the hierarchy in the data is not based on visual aspects. The taxonomic resolution is based on affinity which can be independent of the appearance of the taxa. However, the automatic classifiers base all classification decisions on visual features hence the man-made hierarchy of the data could confuse the classifiers. Fig. \ref{ssbd} gives examples of taxa that belong to the same family or genus but have clear differences in their appearance, e.g., size.

\section{Discussion}

The status assessment of ecosystems is often based on the use of biological indicators that are manually identified by human experts. The manual collection and identification of the data by ecological experts is, however, known to be costly and time-consuming. While recently a growing number of studies explore the enormous potential of genetic identification methods, these are currently not standardized, and thus currently cannot be used to their full potential for legislative biomonitoring purposes \citep[e.g.][]{hering2018}. An interim solution could lie in the use of a computer-based identification system that could be used to simply replace the step of human identification in current biomonitoring while preserving all other steps of the existing process chain. To switch to this novel approach, ecologists must start to put trust in the machine logic. In this work, we compared human expert predictions for physical and image data to those of machine learning methods on image data. 
 
To automate the identification process, we have developed a generic imaging system producing multiple images for each specimen. With our imaging system, we collected a large dataset of benthic freshwater macroinvertebrate images and assigned labels consisting of multiple taxonomic ranks. The classical approach in the computer-based identification has been a flat classification, where the classification is performed at the most specific rank of the taxonomic resolution. In addition to the classical flat approach, we considered also local hierarchical classifiers, namely local per level classifiers and local per parent node classifiers. We selected \acfp{CNN} and \acfp{SVM} as classification methods. We are not aware of any earlier works applying the local hierarchical classifiers based on the taxonomic resolution of invertebrates. We evaluated both automatic classifiers and taxonomic experts using the classification error (CE) at the most specific level and a novel variant of the context sensitivity error (CSE) taking the top-down success into account. We call this variant \acf{LCSE}.
    
We split the image data to produce test sets similar to the ones used in the proficiency test with physical data for taxonomic experts to be able to directly compare machines and human experts. We found that the taxonomic experts obtained the best classification performance when analysing the physical data using a microscope ($\overline{CE}=6.1\%$ and $\overline{LCSE}=2.8\%$) and the worst when using the image data ($\overline{CE}=55.3\%$ and $\overline{LCSE}=35.3\%$). The best automatic classifier was the \ac{CNN} using flat classification approach and the average output of all the images for a specimen as the decision rule to decide the final label ($\overline{CE}=11.4\%$ and $\overline{LCSE}=5.3\%$). This result is well within the range of human experts taking part in the proficiency test. We observed also that, contrary to earlier observations in the literature, the flat classifiers with both \ac{CNN} and \ac{SVM} performed better than the local per parent node hierarchical classifiers. We assume this is because the hierarchy based on the taxonomic resolution does not necessarily correlate with the visual similarity of the taxa. The hierarchical classifiers would be likely more successful if they could first separate the easiest superclasses and then concentrate on more subtle differences within those superclasses. Besides the \ac{CE} and \ac{LCSE} measures, we also investigated the main differences in confusion matrices. The most difficult classes were partially overlapping for machines and experts, but there were some differences as well. Human experts using images preferred to stay at higher ranks of taxonomic hierarchy for difficult taxa while machines were forced to predict the deepest possible level, and thus, ended up predicting wrong species. Unsurprisingly, we observed that \acp{CNN} had trouble identifying the classes with a low amount of training samples.

The test sets in our comparison data were very small to not burden the human participants too much. This naturally makes the results unstable in the sense that few difficult specimens or bad images may affect the results a lot. Therefore, we evaluated the automatic classifiers also on different data splits, where the test sets were considerably larger and also represented the overall taxa distribution. The ranking of the automatic classifiers with respect to the \ac{CE} and \ac{LCSE} measures was similar, while the absolute \ac{CE} and \ac{LCSE} values were much smaller for these larger test sets. Again, forcing automatic classifiers to operate with the logic of human experts, i.e., local per parent node approach, did not improve classification results.

\section{Conclusion}

The main purpose of this paper was to investigate differences in the identification logic of humans and machines. When compared to the existing literature, up to our knowledge this was the first attempt to use a human-like hierarchical classifier for macroinvertebrate image data. With respect to accuracy of identification human taxonomic experts still outperformed the selected automatic methods on the limited set of taxa and specimens used in proficiency tests, but \acp{CNN}' performance was close and fell within the range of typical human experts. With respect to speed, human identification is no match to that of machines' which are are at least X times faster and will be faster still with improved algorithms and increases in computing power. In future studies, we will apply more advanced machine learning techniques, further boost the identification performance on the most rare classes using, e.g., transfer learning and data augmentation, and consider global hierarchical classifiers. 

It is important that ecologists understand and leverage the potential that the high speed and overall good accuracy of automated identification can have on assessments. If applied, these methods will significantly reduce human workload and perform routine identification tasks to a suffieciently accurate degree. Given our results and the fast pace in the field of image recognition, we expect that automatic identification methods can replace human experts in the routine identification of bulk taxa soon, while human experts and genetic methods will still be needed to concentrate on the harder to identify cases. We hope that our results convince doubting ecologists to trust that machine logic can indeed be used to take over a task traditionally done by humans while also increasing their understanding of the main challenges still associated with automatic identification.

\section*{Acknowledgments}
We thank the Academy of Finland for the grants of Ärje (284513, 289076), Tirronen (289076, 289104), Kärkkäinen \\(289076), Meissner (289104), and Raitoharju (288584). We would like to thank CSC for computational resources.

\appendix
\section{}

See Tables~\ref{numbers} and \ref{features}. \\

\section*{References}
\bibliographystyle{plainnat}
\bibliography{references}

%\clearpage
%\section*{Appendix}

%\begin{landscape}
\begin{table*}[h]
\scriptsize
\begin{center}
\begin{tabular}{|l|l|l|l|l|c|c|}
\hline
Taxa&	Species&	Genus&	Family&	Order&	\#specimens&	\#images\\
\hline
\textit{Elmis aenea}&	\textit{Elmis aenea}&	\textit{Elmis}&	\textit{Elmidae}&	\textit{Coleoptera}&	648&	32398\\
\textit{Limnius volckmari}&	\textit{Limnius volckmari}&	\textit{Limnius}&	\textit{Elmidae}&	\textit{Coleoptera}&	314&	15621\\
\textit{Oulimnius tuberculatus}&	\textit{Oulimnius tuberculatus}&	\textit{Oulimnius}&	\textit{Elmidae}&	\textit{Coleoptera}&	335&	16674\\
\textit{Hydraena sp.}&	\textit{-}&	\textit{Hydraena}&	\textit{Hydraenidae}&	\textit{Coleoptera}&	198&	9900\\
\textit{Simuliidae}&	\textit{-}&	\textit{-}&	\textit{Simuliidae}&	\textit{Diptera}&	887&	44240\\
\textit{Ameletus inopinatus}&	\textit{Ameletus inopinatus}&	\textit{Ameletus}&	\textit{Ameletidae}&	\textit{Ephemeroptera}&	127&	6346\\
\textit{Baetis rhodani}&	\textit{Baetis rhodani}&	\textit{Baetis}&	\textit{Baetidae}&	\textit{Ephemeroptera}&	404&	19829\\
\textit{Baetis vernus group}&	\textit{Baetis vernus}&	\textit{Baetis}&	\textit{Baetidae}&	\textit{Ephemeroptera}&	176&	8588\\
\textit{Ephemerella aurivillii}&	\textit{Ephemerella aurivillii}&	\textit{Ephemerella}&	\textit{Ephemerellidae}&	\textit{Ephemeroptera}&	356&	16458\\
\textit{Ephemerella mucronata}&	\textit{Ephemerella mucronata}&	\textit{Ephemerella}&	\textit{Ephemerellidae}&	\textit{Ephemeroptera}&	304&	15175\\
\textit{Heptagenia sulphurea}&	\textit{Heptagenia sulphurea}&	\textit{Heptagenia}&	\textit{Heptageniidae}&	\textit{Ephemeroptera}&	438&	21502\\
\textit{Kageronia fuscogrisea}&	\textit{Kageronia fuscogrisea}&	\textit{Kageronia}&	\textit{Heptageniidae}&	\textit{Ephemeroptera}&	222&	10826\\
\textit{Leptophlebia sp.}&	\textit{-}&	\textit{Leptophlebia}&	\textit{Leptophlebiidae}&	\textit{Ephemeroptera}&	412&	20366\\
\textit{Sialis sp.}&	\textit{-}&	\textit{Sialis}&	\textit{Sialidae}&	\textit{Megaloptera}&	26&	1162\\
\textit{Capnopsis schilleri}&	\textit{Capnopsis schilleri}&	\textit{Capnopsis}&	\textit{Capniidae}&	\textit{Plecoptera}&	21&	1050\\
\textit{Leuctra nigra}&	\textit{Leuctra nigra}&	\textit{Leuctra}&	\textit{Leuctridae}&	\textit{Plecoptera}&	27&	1350\\
\textit{Leuctra sp.}&	\textit{-}&	\textit{Leuctra}&	\textit{Leuctridae}&	\textit{Plecoptera}&	298&	14899\\
\textit{Amphinemura borealis}&	\textit{Amphinemura borealis}&	\textit{Amphinemura}&	\textit{Nemouridae}&	\textit{Plecoptera}&	322&	16100\\
\textit{Nemoura cinerea}&	\textit{Nemoura cinerea}&	\textit{Nemoura}&	\textit{Nemouridae}&	\textit{Plecoptera}&	16&	800\\
\textit{Nemoura sp.}&	\textit{-}&	\textit{Nemoura}&	\textit{Nemouridae}&	\textit{Plecoptera}&	187&	9314\\
\textit{Protonemura sp.}&	\textit{-}&	\textit{Protonemura}&	\textit{Nemouridae}&	\textit{Plecoptera}&	100&	4908\\
\textit{Diura sp.}&	\textit{-}&	\textit{Diura}&	\textit{Perlodiae}&	\textit{Plecoptera}&	98&	4427\\
\textit{Isoperla sp.}&	\textit{-}&	\textit{Isoperla}&	\textit{Perlodiae}&	\textit{Plecoptera}&	243&	12148\\
\textit{Taeniopteryx nebulosa}&	\textit{Taeniopteryx nebulosa}&	\textit{Taeniopteryx}&	\textit{Taenioptegyridae}&	\textit{Plecoptera}&	331&	16325\\
\textit{Micrasema gelidum}&	\textit{Micrasema gelidum}&	\textit{Micrasema}&	\textit{Brachycentridae}&	\textit{Trichoptera}&	233&	11528\\
\textit{Micrasema setiferum}&	\textit{Micrasema setiferum}&	\textit{Micrasema}&	\textit{Brachycentridae}&	\textit{Trichoptera}&	323&	13819\\
\textit{Agapetus sp.}&	\textit{-}&	\textit{Agapetus}&	\textit{Glossosomatidae}&	\textit{Trichoptera}&	290&	14387\\
\textit{Silo pallipes}&	\textit{Silo pallipes}&	\textit{Silo}&	\textit{Goeridae}&	\textit{Trichoptera}&	56&	2658\\
\textit{Hydropsyche pellucidula}&	\textit{Hydropsyche pellucidula}&	\textit{Hydropsyche}&	\textit{Hydropsychidae}&	\textit{Trichoptera}&	192&	6513\\
\textit{Hydropsyche saxonica}&	\textit{Hydropsyche saxonica}&	\textit{Hydropsyche}&	\textit{Hydropsychidae}&	\textit{Trichoptera}&	17&	490\\
\textit{Hydropsyche siltalai}&	\textit{Hydropsyche siltalai}&	\textit{Hydropsyche}&	\textit{Hydropsychidae}&	\textit{Trichoptera}&	395&	19456\\
\textit{Oxyethira sp.}&	\textit{-}&	\textit{Oxyethira}&	\textit{Hydroptilidae}&	\textit{Trichoptera}&	218&	10381\\
\textit{Lepidostoma hirtum}&	\textit{Lepidostoma hirtum}&	\textit{Lepidostoma}&	\textit{Lepidostomatidae}&	\textit{Trichoptera}&	267&	10982\\
\textit{Neureclipsis bimaculata}&	\textit{Neureclipsis bimaculata}&	\textit{Neureclipsis}&	\textit{Polycentropodidae}&	\textit{Trichoptera}&	477&	23721\\
\textit{Plectrocnemia sp.}&	\textit{-}&	\textit{Plectrocnemia}&	\textit{Polycentropodidae}&	\textit{Trichoptera}&	63&	3015\\
\textit{Polycentropus flavomaculatus}&	\textit{Polycentropus flavomaculatus}&	\textit{Polycentropus}&	\textit{Polycentropodidae}&	\textit{Trichoptera}&	224&	11005\\
\textit{Polycentropus irroratus}&	\textit{Polycentropus irroratus}&	\textit{Polycentropus}&	\textit{Polycentropodidae}&	\textit{Trichoptera}&	59&	2917\\
\textit{Rhyacophila nubila}&	\textit{Rhyacophila nubila}&	\textit{Rhycophila}&	\textit{Rhyacophilidae}&	\textit{Trichoptera}&	177&	6993\\
\textit{Sphaerium sp.}&	\textit{-}&	\textit{Sphaerium}&	\textit{Sphaeridae}&	\textit{Veneroida}&	150&	1733\\
\hline
\end{tabular}
\caption{Taxonomic resolution of the multiple image data and the numbers of specimens and images per taxa. Taxa included in the proficiency test for human experts but not included in the image data were \textit{Brachyptera risi}, \textit{Cloeon sp.}, \textit{Cloeon diptera group}, \textit{Cloeon inscriptum}, \textit{Cloeon simile}, \textit{Helobdella stagnalis}, and \textit{Tinodes waeneri}.}
\label{numbers}
\end{center}
\end{table*}
%\end{landscape}

\begin{table*}[h]
	\begin{center}
		\begin{tabular}{|l|l|}
		\hline
		Geometric features & RGB and grey scale features\\
		\hline
		Area & Mean\\
		Center of mass & Standard deviation\\
		\tabitem X and Y coordinates & Mode\\
		Perimeter & Minimum \\
		Bounding rectangle & Maximum\\
		\tabitem Width and Height & Center of mass\\
		\tabitem X and Y coordinates of the upper left corner & \tabitem X and Y coordinates\\
		Fit ellipse & Integrated density\\
		\tabitem Major and Minor axis & Median\\
		\tabitem Angle & Skewness\\
		\tabitem X and Y coordinates of the center & Kurtosis\\
		Circularity & \\
		Aspect ratio & \\
		Roundness & \\
		Solidity & \\
		Feret's diameter & \\
		\tabitem Length & \\
		\tabitem Angle & \\
		\tabitem Minimum caliper length & \\
		\tabitem X and Y starting coordinates & \\
		\hline
	\end{tabular}
	\end{center}
	\caption{Features used for \ac{SVM} classification.}
	\label{features}
\end{table*}

\end{document}